\documentclass{article}

\usepackage{xcolor}
\definecolor{dullorange}{RGB}{204,119,34}

\usepackage{times}
\usepackage{soul}
\usepackage{url}
\usepackage[hidelinks]{hyperref}
\usepackage[utf8]{inputenc}
\usepackage[small]{caption}

\PassOptionsToPackage{numbers, compress}{natbib}

\usepackage{amsmath}
\usepackage{amsthm}
\usepackage{booktabs}
\usepackage{algorithm}
\usepackage{algorithmic}
\usepackage[switch]{lineno}
\usepackage{xcolor}
\usepackage{natbib}
\usepackage{float} 

\usepackage[T1]{fontenc}    
\usepackage{hyperref}       
\usepackage{url}            
\usepackage{booktabs}       
\usepackage{amsfonts}       
\usepackage{nicefrac}       
\usepackage{microtype}      
\usepackage{lipsum}		
\usepackage{dirtytalk}
\usepackage{xcolor}
\usepackage{doi}
\usepackage{booktabs}
\usepackage{siunitx}
\usepackage[table]{xcolor}
\usepackage{colortbl}
\usepackage{calc}

\usepackage{multirow}

\usepackage{xcolor}

\usepackage{tcolorbox}
\tcbuselibrary{breakable}

\usepackage{tabularx}
\usepackage{booktabs}  
\usepackage{xcolor}    
\usepackage{array}     
\usepackage{lipsum}    
\usepackage{fancyvrb}
\usepackage{fvextra} 
\usepackage{calc}    
\usepackage{graphicx}
\usepackage{longtable}
\usepackage{adjustbox}
\usepackage{subcaption} 
\PassOptionsToPackage{numbers, compress}{natbib}

\usepackage[sglblindworkshop, final]{neurips_2025}
\workshoptitle{Multi-Turn Interactions in Large Language Models}

\usepackage[utf8]{inputenc} 
\usepackage[T1]{fontenc}    
\usepackage{hyperref}       
\usepackage{url}            
\usepackage{booktabs}       
\usepackage{amsfonts}       
\usepackage{nicefrac}       
\usepackage{microtype}      
\usepackage{xcolor}         
\usepackage{graphicx} 
\title{MAREval: A Multi-Agent Framework for Evaluating Natural Language Recommendation Explanations}

\author{%
  Reza Yousefi Maragheh$^{*\dagger}$\quad
  Jayesh Uddhav Kudase$^{*}$\quad
  Aysenur Inan$^{*}$\\
  \textbf{Ramin Giahi$^{*}$\quad
  Kai Zhao$^{*}$\quad
  Parth Parikh\quad
  Jianpeng Xu}\\
  \textbf{Jason Cho\quad
  Evren Korpeoglu\quad
  Sushant Kumar}\\[4pt]
  Walmart Global Tech, Sunnyvale, CA, USA
}

\begin{document}

\maketitle
{\renewcommand{\thefootnote}{$\fnsymbol{footnote}$}%
 \footnotetext[1]{Equal contribution.}%
 \footnotetext[2]{Corresponding author: \texttt{reza.yousefimaragheh@walmart.com}}}
\begin{abstract}
Evaluating natural language explanations in recommender systems is essential for fostering user trust, transparency, and engagement. However, existing evaluation approaches like human evaluations, while accurate, are resource-intensive and impractical at the scale required by modern recommendation platforms. Also, automated methods using single-agent LLMs suffer from prompt sensitivity and inconsistent outputs. To address these challenges, we propose MAREval, a structured multi-agent framework for evaluating recommendation explanations using large language models.  MAREval orchestrates (i) a planner agent that uses a novel Chain of Debate (CoD) prompting strategy to coordinate agent roles and enforce logically consistent evaluation plans; (ii) a moderator agent that regulates discussions by mitigating prompt drift; and (iii) an arbitrator agent that aggregates outputs from multiple evaluation rounds and (iv) a Monte Carlo sampling method, improving robustness and alignment with human judgment. We conduct comprehensive evaluations on both public (TopicalChat) and proprietary recommendation datasets, demonstrating that MAREval outperforms state-of-the-art baselines. Comprehensive experiments on a public benchmark and a proprietary e‑commerce dataset show that MAREval improves alignment with human judgments over strong single‑ and multi‑agent baselines. Stability analyses indicate substantially lower variability across repeated trials. In a large human‑annotation gate, MAREval meets production quality thresholds where prior evaluators fall short, and online A/B testing demonstrates statistically significant improvements in engagement and revenue metrics. These results establish MAREval as a scalable and reliable solution for human-aligned evaluation of recommendation explanations in real-world systems.
\end{abstract}

\section{Introduction}

\begin{figure*}
  \includegraphics[width=0.95\textwidth]{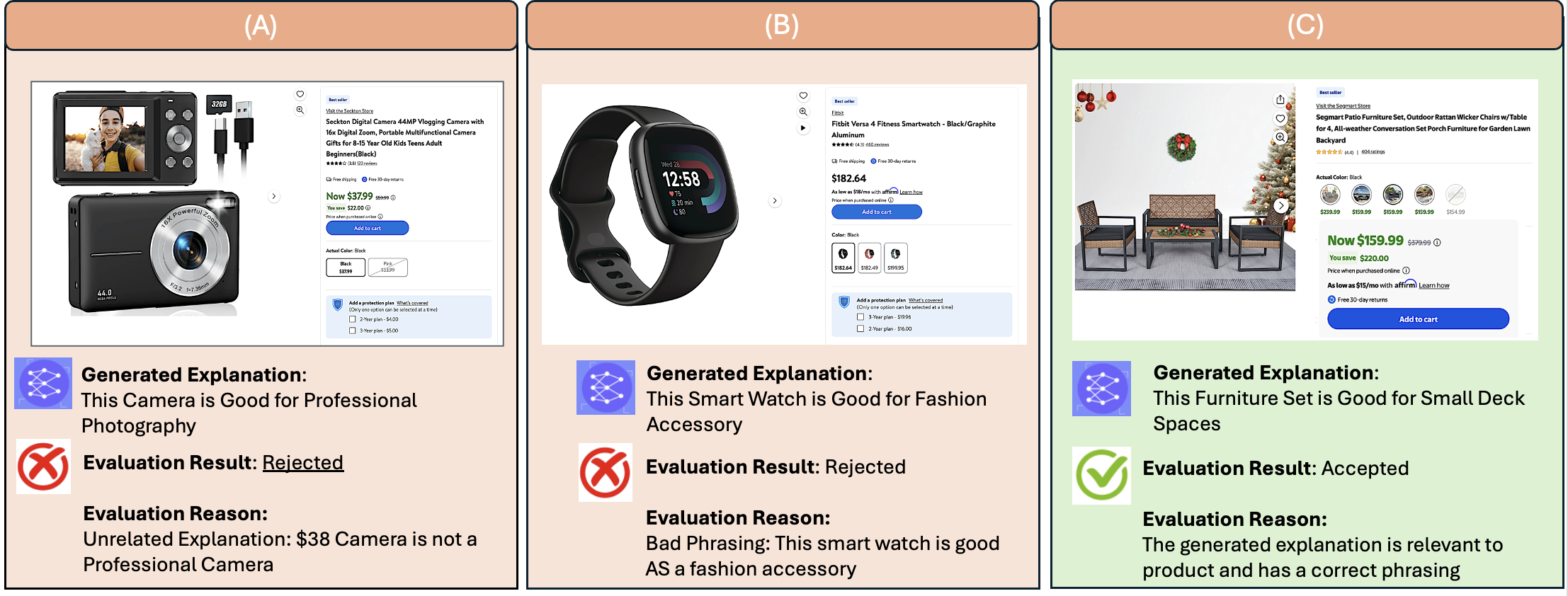}
  \centering
  \caption{Natural Language Evaluation of Generated Recommendation Explanations for eCommerce Products.}
  \label{fig:enter-label}
\end{figure*}

Millions of users daily interact with recommendation systems to discover tailored products, movies, or music. System effectiveness hinges not just on recommendations but on accompanying explanations that build trust and guide informed decisions \citep{maragheh2025future,zhang2020explainable, balog2020measuring, chen2021generate}. Poor-quality explanations with misleading or unlear message lead to user frustration and decreased platform reliability \citep{ji2023towards, wang2023large}. Figure~\ref{fig:enter-label} illustrates this: case (A) shows a budget camera misleadingly described for professional photography, case (B) presents a smartwatch with poorly phrased explanation, while only case (C) demonstrates ideal explanation quality. Traditional recommendation explanation evaluation relies on expert annotators manually assessing explanations for relevance, phrasing, and informativeness. While ensuring quality, this approach is expensive, time-consuming, and impractical for large datasets \citep{lin2004rouge, ji2023towards}.

Recent advances in large language models have led to automated evaluation methods \cite{zheng2024judging}. This is achieved through single model judges or sophisticated strategies like chain-of-thought prompting \cite{wei2022chain, wang2022self, chowdhery2023palm}, self-refinement \cite{madaan2023self, zeng2025aries}, and multi-agent frameworks \cite{chan2023chateval}. However, existing solutions suffer critical issues: single agent evaluators exhibit bias and inconsistency \cite{liu2023llmbias}, multi-agent frameworks lack structured coordination, leading to noisy debates \cite{wu2023autogen}. These frameworks may also lead to evaluation variability, which means that identical explanations may receive different scores in the same framework \cite{duan2024llms}.

We introduce MAREval, a structured multi-agent evaluation framework enhancing recommendation explanation evaluation through orchestrated planning, structured debate, and statistical refinement. MAREval leverages a novel ``Chain of Debate'' (CoD) mechanism, a planning-first prompting strategy structuring evaluation in multi-agent frameworks. Unlike traditional multi-agent systems where agents interact in an ad hoc or loosely coordinated fashion, MAREval introduces a planner agent that first analyzes the overall flow and then generates synchronized, role-specific plans for each agent involved in the discussion. This prestructuring promotes logical consistency in the dialogue and encourages complementary reasoning between agents. This ensures less evaluation output consistency with ground truth data.

A dedicated moderator agent orchestrates the conversation flow, distilling the evaluations into concise summaries that capture agreement and contention points. This grounds subsequent evaluations in shared context, preventing redundancy and noise. In doing so, the moderator serves as a dynamic memory mechanism, enabling more effective information sharing without overloading the prompt context. This structured regulation contributes to more stable and interpretable evaluations.

Finally, MAREval employs an arbitrator agent to judge outcomes across multiple debate rounds and to counteract variability in LLM outputs. In addition to this arbitration agent, and given the inherent randomness in language model generation MAREval uses a Monte Carlo sampling strategy that aggregates multiple independently sampled runs of the agent debates to arrive at a final decision. Together with arbitration, the Monte Carlo sampling improve alignment with human annotations and reduce the likelihood of outlier judgments. Extensive experiments demonstrate MAREval achieves higher consistency with human evaluations than existing methods. Online A/B tests validate effectiveness, revealing measurable improvements in user engagement and conversion metrics.

\section{Related Works}\label{sec:background}

With LLM advancements, researchers utilize them for natural language evaluation tasks \cite{zheng2024judging, maragheh2023llm, li2023alpacaeval, kim2023prometheus}. \citet{du2023improving} systematically showed that using LLMs in debate settings improves reasoning capabilities and output quality. \citet{liang2023encouraging} identifies the "Degeneration of Thought" problem in frameworks like Self-Refine \cite{madaan2023self}, where LLMs struggle to self-refine already generated evaluations. This led to the "Multi-Agent Debate" (MAD) framework, where agents discuss answers to generate refined responses \citep{liang2023encouraging}.

After demonstrating multi-agent framework efficacy, researchers applied them to evaluation tasks. \citet{li2023prd} proposes Peer Discussion, utilizing two LLM agents to discuss evaluation tasks and reach agreement. Under CHATEVAL \cite{chan2023chateval}, evaluators collectively conduct evaluation tasks with different communication strategies. Agentic pipelines have been applied to recommendation domains \cite{peng2025survey}, focusing on: (i) context-aware recommendations \cite{wang2024macrec, zhao2024let, shu2024rah, giahi2025vl}, (ii) enhancing recommendations through interactive user engagement \cite{friedman2023leveraging, fang2024multi, zhang2024prospect}, (iii) simulating user behavior \cite{zhu2024llm, yoon2024evaluating}, and (iv) recommendation explanation \cite{chen2025carts}.

While prior works like CHATEVAL, Peer Discussion, and MAD demonstrated multi-agent debate promise, they often struggle with prompt drift, redundancy, and inconsistency in evaluation outputs. MAREval introduces a novel Chain of Debate planning mechanism that explicitly coordinates agent roles and reasoning sequences. Additionally, MAREval uniquely integrates a moderator agent and an arbitration agent for communication regulation and evaluation aggregation respectively. This significantly reduces variability and improving human judgment alignment. Also, through a Monte-Carlo sampling method MAREval stabilizes outlier output and improves consistency with ground truth judgments. These innovations make MAREval the first end-to-end multi-agent evaluation framework optimized for both general dialogue tasks and recommendation-specific explanation evaluation.

\section{Methodology}\label{sec:method}

\begin{figure*}[t]
  \centering
  \includegraphics[width=\textwidth]
  {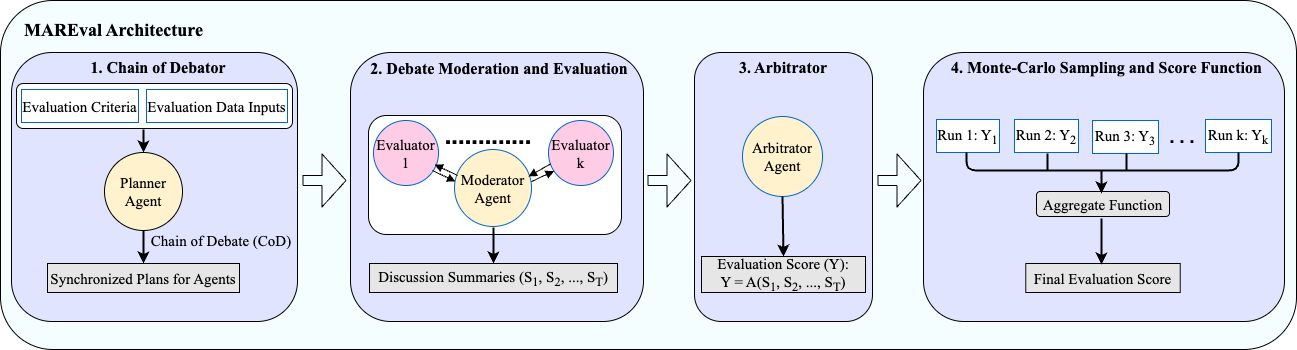}
  \caption{MAREval Framework. Under MAREval, (1) the Planner Agent is prompted to generate synchronized chain of steps for individual agents of the pipelines given the overall flow design of the pipeline. (2) The discussion among the evaluator agents is moderated by a moderator agent to better conduct the evaluation task. (3) The result of discussions is then passed to the Arbitrator Agent to generate the final ruling on the evaluation task. (4) Monte Carlo Sampling method applied to aggregate different rounds of MAREval block}
  \label{evals}
\end{figure*}

In this section, we review the major elements of MAREval including Chain of Debate (CoD) used for planning, Context Moderation (MoD) realized by moderation agent, arbitration, and Multi-Agent Monte-Carlo Sampling (MAMCS). We introduce formal notation and discuss the high-level intuition and logic for each component. We show a schematic view of the evaluation task in MAREval framework (see Figure \ref{evals}).

\subsection{Chain of Debate} 

Chain of Debate (CoD) extends Chain of Thought prompting to multi-agent settings by introducing a planner agent that generates synchronized, step-wise plans for all agents while considering the complete framework flow. Unlike independent agent planning, CoD ensures coordinated collaboration where each agent's steps depend on inter-agent information flow.

The planner receives evaluation criteria, contextual information, and framework flow description to generate comprehensive plans for evaluator agents, moderator, and arbitrator. This synchronized planning promotes coherent step-by-step coordination across all agents, addressing the challenge of inconsistent and independent agents' plans in multi-agent environments.

Formally, let $C$ be evaluation criteria (e.g., $\{\text{relevance}, \text{phrasing}\}$), $I$ be contextual information, $F$ be the multi-agent pipeline flow description, and $T$ be total evaluation rounds. Denote $m$ evaluator agents as $\mathcal{A} = \{\alpha_1, \dots, \alpha_m\}$, moderator agent as $\alpha_M$, and arbitrator agent as $\alpha_A$.

CoD is defined as: $\text{Planner}(C, I, F) = \Bigl(\{P^e_t\}_{t=1}^T,\{P^m_t\}_{t=1}^T,P^a\Bigr)$, where $P^e_t$ is the evaluator agent plan for round $t$, $P^m_t$ is the moderator plan for round $t$, and $P^a$ is the arbitrator plan.

During execution, the planner dispatches appropriate plans to respective agents in each round. Agents produce intermediate outputs following their instructions while information flows according to $F$. Since the plans $\{P^e_t\}_{t=1}^T$, $\{P^m_t\}_{t=1}^T$, and $P^a$ are generated in a synchronized manner with full knowledge of $(C, I, F)$, CoD ensures coherent, step-by-step coordination across all agents. Although we present CoD in the context of an evaluation task, it can be extended naturally to other NLG workflows by adapting the agent roles accordingly.

\subsection{Context Moderation and Arbitration}

The \emph{moderator agent} governs information flow among evaluator agents by distilling key discussion points from previous rounds. Inspired by human-style debates and recent studies on memory mechanisms in multi-agent frameworks \cite{chen2023autoagents}, the moderator manages context without excessive verbosity or noise, storing essential points for subsequent rounds.

\emph{Evaluator agents} apply plan segments to ongoing discussions, critiquing and refining judgments while flagging errors or hallucinations. At debate round $t$, the $j$-th evaluator generates evaluation $e_{j,t}$ based on plan $P^e_t$ and discussion history $H_{t-1}$:
$$e_{j,t} = f(P^e_t, H_{t-1}), \quad H_{t-1} = \{S_1, S_2, \ldots, S_{t-1}\},$$
where $S_\tau$ is the moderator summary after round $\tau$.

The moderator combines partial evaluations $e_{1,t}, \ldots, e_{m,t}$ into noise-reduced summary $S_t$, capturing essential disagreements and refinements while compressing repetitive details:
$$S_{t} = M(e_{1,t}, e_{2,t}, \ldots, e_{m,t} | S_{t-1}).$$

While we focus on textual summarization, the moderation mechanism can be generalized. An embedding-based approach might maintain vector representations of evaluations, where $M(\cdot)$ operates on both textual inputs and embeddings:
$$(S_{t}, Z_{t}) = M(\{ e_{i,t} \}_{i=1}^m, S_{t-1}, Z_{t-1}),$$
with $Z_{t-1}$ representing internal memory. This enables similarity detection for redundancy filtering, argument clustering, and noise reduction. In the final round, the moderator provides conclusive summary $S_{T}$ to the arbitrator for final decision-making.

After $T$ rounds of debate and moderation, the \emph{arbitrator agent} receives the final moderator output $S_T$ to produce the ultimate evaluation outcome $Y$. The arbitrator employs structured decision-making to enhance consistency and reduce variability in final evaluations. For scalar evaluation scores, the arbitrator aggregates the final summary $S_T$ with the planner's arbitration plan $P^a$, evaluation criteria $C$, and instance context $I$:
$$Y = g(P^a, C, I, S_T),$$
where $g(\cdot)$ converts the moderated summary into the conclusive evaluation outcome. The framework can be generalized to incorporate richer information. If the moderator maintains internal representation $Z_T$ (e.g., neural encoding of the discussion), the arbitrator can leverage both textual synopsis and semantic relationships:
$$Y = g(P^a, C, I, S_T, Z_T).$$

\subsection{Monte Carlo Sampling and Score Function}

Language models' probabilistic token sampling introduces high variability in multi-agent settings, where single-agent variance compounds in final outputs. Inspired by \cite{liu2023gpteval}, we propose Monte Carlo sampling for multi-agent evaluation frameworks to mitigate outlier results through multiple pipeline runs and result aggregation. To the best of our knowledge, the effect of sampling methods in the performance of multi-agent evaluation frameworks has not been studied. 

For models providing token-level probabilities, we compute expected outcomes directly. Let $Y = \{y_1, \dots, y_n\}$ be possible final tokens or labels, with $P(y_i | P^a, C, I, S_T)$ as the probability of outputting $y_i$. The Monte Carlo weighted outcome is:
\begin{equation}
    y^{MC} = \sum_{i=1}^n y_i \times P(y_i | P^a, C, I, S_T).
    \label{eq:MC_token}
\end{equation}

For black-box models without probability access, we apply pipeline-level sampling by running the complete multi-agent system $N$ times independently. The final outcome from run $k$ is $Y_k = \{g(P^a, C, I, S_T)\}_k$, aggregated via arithmetic mean:
\begin{equation}
    \begin{cases}
        Y_k = \{g(P^a, C, I, S_T)\}_k, & k=1,\dots,N,\\[6pt]
        Y_{\text{final}}^N = \frac{1}{N}\sum_{k=1}^N Y_k.
    \end{cases}
    \label{eq:MC_pipeline}
\end{equation}

Both strategies address LLM sampling randomness, reducing reliance on single trajectories to achieve more stable evaluation outcomes across multi-agent system runs (See Section \ref{std:analysis}). Refer to Appendix \ref{app:output}, \ref{app:sample} and \ref{app:input} for MAREval's sample outputs and inputs.

\section{Experiments}\label{sec:experiment}
\subsection{Implementation Details and Datasets}

We perform evaluation experiments on two datasets: recommendation explanation evaluation on a proprietary e-commerce dataset and response generation evaluation on the public TopicalChat dataset \cite{zhong2022towards, gopalakrishnan2023topical}: (i) \textbf{Proprietary E-commerce Dataset}: Our Proprietary E-commerce (PE) dataset contains 108 recommendation explanation phrases for target items, describing product "use cases" (e.g., "gaming" for laptops, "heart-rate monitoring" for smartwatches). Phrases are manually evaluated by product domain experts on two binary criteria: relevance to the item and phrasing quality (grammatical correctness). We compare MAREval and benchmark model results with these expert annotations. (ii) \textbf{TopicalChat Dataset}: TopicalChat \cite{gopalakrishnan2023topical} contains responses generated for chat histories with contextual considerations. The dataset includes manual annotations across six criteria: naturalness, understandability, groundedness, engagingness, maintaining context, and overall quality. While not directly containing product recommendations, TopicalChat provides conversational contexts mirroring recommendation scenarios requiring personalized responses based on user preferences and history. All of the results are based on a gpt-3.5-turbo model.

\subsection{Metrics and Benchmarks}

We evaluate using Cohen's Kappa ($\kappa$) and Pearson's correlation ($\rho$) for binary outcomes, and Pearson ($\rho$) and Kendall-Tau ($\tau$) correlations for integer range outcomes \cite{chan2023chateval, liu2023gpteval}. These metrics are standard for measuring agreement between evaluation frameworks and manual annotators.

We compare against the following benchmark models: (i) \textit{Vanilla-gpt}: Simple instruction prompting asking the LLM to conduct evaluation based on given criteria. (ii) \textit{Chain-of-Thought} \cite{wei2022chain}: CoT prompting where the LLM generates step-by-step evaluation plans before conducting assessment. (iii) \textit{Self-Refine} \cite{madaan2023self}: Response refinement technique where a single LLM generates then refines its evaluation response. (iv) \textit{G-Eval} \cite{liu2023gpteval}: Combines CoT prompting with score function sampling, prompting GPT multiple times and averaging results. (v) \textit{ChatEval} \cite{chan2023chateval}: State-of-the-art multi-agent framework using sequential evaluator chats with majority vote for final outcomes.

\begin{table*}[!t]
\centering
\small 
\caption{Performance of different frameworks across TopicalChat and PE datasets. The variants of MAREval are compared to five benchmark models. The table shows Pearson Correlation ($\rho$), Kendall-Tau ($\tau$), and Cohen's kappa ($\kappa$) metrics for different evaluation criteria.}
\resizebox{\textwidth}{!}{%
\begin{tabular}{
  l
  S[table-format=1.3]
  S[table-format=1.3]
  S[table-format=1.3]
  S[table-format=1.3]
  S[table-format=1.3]
  S[table-format=1.3]
  S[table-format=1.3]
  S[table-format=1.3]
}
\toprule
&\multicolumn{8}{c}{\textbf{TopicalChat (TC)}} \\
\midrule
Metrics & \multicolumn{2}{c}{Naturalness} & \multicolumn{2}{c}{Engagingness} & \multicolumn{2}{c}{Maintains Context} & \multicolumn{2}{c}{Understandability} \\
\cmidrule(r){2-3} \cmidrule(lr){4-5} \cmidrule(l){6-7} \cmidrule(l){8-9}
& {$\rho$} & {$\tau$} &  {$\rho$} & {$\tau$} &  {$\rho$} & {$\tau$} & {$\rho$} & {$\kappa$} \\
\midrule
CoT & 0.351 & 0.317 & 0.309 & 0.292 & 0.490 & 0.469 & 0.270 & 0.215 \\
Self-Refine & 0.221 & 0.200 & \textbf{0.355} & \underline{0.325} & 0.408 & 0.384 & 0.167 & 0.144 \\
G-eval & \underline{0.367} & \underline{0.332} & 0.284 & 0.270 & 0.493 & 0.470 & 0.281 & 0.209 \\
Vanilla & 0.291 & 0.266 & 0.203 & 0.193 & \underline{0.522} & \underline{0.499} & \textbf{0.336} & \textbf{0.304} \\
ChatEval & 0.290 & 0.269 & 0.289 & 0.268 & 0.482 & 0.466 & 0.253 & \underline{0.228} \\
\midrule
MAREval & \textbf{0.389} & \textbf{0.350} & \underline{0.341} & \textbf{0.327} & \textbf{0.524} & \textbf{0.508} & \underline{0.284} & 0.181 \\
\midrule
\midrule
& \multicolumn{4}{c|}{\textbf{TopicalChat (TC)}} & \multicolumn{4}{c}{\textbf{PE Dataset}} \\
\midrule
Metrics & \multicolumn{2}{c}{Groundedness (TC)} & \multicolumn{2}{c|}{Overall Chat Quality (TC)} & \multicolumn{2}{c}{Relevance (PE)} & \multicolumn{2}{c}{Phraseness (PE)} \\
\cmidrule(r){2-3} \cmidrule(l){4-5} \cmidrule(l){6-7} \cmidrule(l){8-9}
& {$\rho$} & {$\kappa$} & {$\rho$} & {$\tau$} & \multicolumn{1}{|c}{$\rho$} & {$\kappa$} & {$\rho$} & {$\kappa$} \\
\midrule
\multicolumn{1}{l}{CoT} & \multicolumn{1}{S[table-format=1.3]}{0.485} & \multicolumn{1}{S[table-format=1.3]}{0.485} & \multicolumn{1}{S[table-format=1.3]}{0.442} & \multicolumn{1}{S[table-format=1.3]}{0.388} & \multicolumn{1}{|S[table-format=1.3]}{0.084} & \multicolumn{1}{S[table-format=1.3]}{0.062} & \multicolumn{1}{S[table-format=1.3]}{0.040} & \multicolumn{1}{S[table-format=1.3]}{0.029} \\
\multicolumn{1}{l}{Self-Refine} & \multicolumn{1}{S[table-format=1.3]}{0.412} & \multicolumn{1}{S[table-format=1.3]}{0.412} & \multicolumn{1}{S[table-format=1.3]}{0.469} & \multicolumn{1}{S[table-format=1.3]}{0.401} & \multicolumn{1}{|S[table-format=1.3]}{0.075} & \multicolumn{1}{S[table-format=1.3]}{0.046} & \multicolumn{1}{S[table-format=1.3]}{0.026} & \multicolumn{1}{S[table-format=1.3]}{0.012} \\
\multicolumn{1}{l}{G-eval} & \multicolumn{1}{S[table-format=1.3]}{\underline{0.521}} & \multicolumn{1}{S[table-format=1.3]}{\underline{0.502}} & \multicolumn{1}{S[table-format=1.3]}{0.487} & \multicolumn{1}{S[table-format=1.3]}{0.431} & \multicolumn{1}{|S[table-format=1.3]}{0.070} & \multicolumn{1}{S[table-format=1.3]}{0.045} & \multicolumn{1}{S[table-format=1.3]}{-0.035} & \multicolumn{1}{S[table-format=1.3]}{-0.023} \\
\multicolumn{1}{l}{Vanilla} & \multicolumn{1}{S[table-format=1.3]}{0.455} & \multicolumn{1}{S[table-format=1.3]}{0.455} & \multicolumn{1}{S[table-format=1.3]}{\underline{0.505}} & \multicolumn{1}{S[table-format=1.3]}{\underline{0.438}} & \multicolumn{1}{|S[table-format=1.3]}{0.102} & \multicolumn{1}{S[table-format=1.3]}{0.058} & \multicolumn{1}{S[table-format=1.3]}{0.018} & \multicolumn{1}{S[table-format=1.3]}{0.011} \\

\multicolumn{1}{l}{ChatEval} & \multicolumn{1}{S[table-format=1.3]}{0.477} & \multicolumn{1}{S[table-format=1.3]}{0.439} & \multicolumn{1}{S[table-format=1.3]}{0.455} & \multicolumn{1}{S[table-format=1.3]}{0.390} & \multicolumn{1}{|S[table-format=1.3]}{\underline{0.238}} & \multicolumn{1}{S[table-format=1.3]}{\underline{0.225}} & \multicolumn{1}{S[table-format=1.3]}{\underline{0.203}} & \multicolumn{1}{S[table-format=1.3]}{\underline{0.196}} \\
\midrule
\multicolumn{1}{l}{MAREval} & \multicolumn{1}{S[table-format=1.3]}{\textbf{0.553}} & \multicolumn{1}{S[table-format=1.3]}{\textbf{0.536}} & \multicolumn{1}{S[table-format=1.3]}{\textbf{0.548}} & \multicolumn{1}{S[table-format=1.3]}{\textbf{0.478}} & \multicolumn{1}{|S[table-format=1.3]}{\textbf{0.338}} & \multicolumn{1}{S[table-format=1.3]}{\textbf{0.334}} & \multicolumn{1}{S[table-format=1.3]}{\textbf{0.216}} & \multicolumn{1}{S[table-format=1.3]}{\textbf{0.208}} \\
\bottomrule
\end{tabular}
}
\label{table:combined-performance-table1}
\end{table*}

\subsection{Overall Performance}

In this section, we investigate the performance of different frameworks with human annotators. Results are reported in Tables \ref{table:combined-performance-table1}. Vanilla and G-eval demonstrates strong performance on the TopicalChat dataset but struggles significantly with the PE (recommendation) dataset. Vanilla achieves the second-best performance in Overall Chat Quality ($\rho = 0.505$, $\tau = 0.438$), while G-eval follows closely ($\rho = 0.487$, $\tau = 0.431$) for TopicalChat. 
However, both models show dramatically reduced effectiveness when evaluating PE metrics
This pattern suggests the lack of robustness for these two models when performing on recommendation evaluation tasks.

Self-Refine and CoT show middling performance across both datasets. Self-Refine ranks fourth in Overall Chat Quality ($\rho = 0.469$, $\tau = 0.401$) but falls to the bottom tier for PE metrics. Similarly, CoT shows the weakest performance in TC Overall Chat Quality among all models tested ($\rho = 0.442$, $\tau = 0.388$) and only marginally better results than Self-Refine and G-eval on PE metrics. 
This is due to the effect of the \say{degeneration of thought} \cite{du2023improving} with Self-Refine and CoT, where a single agent cannot correct their opinion once biased from the beginning.

ChatEval show better performance on TopicalChat's Overall Chat Quality but emerging as the clear second-best performer on both PE dataset metrics. ChatEval demonstrates considerably better cross-domain capability than other benchmarks. This suggest ChatEval and multi-agent systems in general can achieve robust performance across evaluation tasks native to recommendation setting and they can safely be employed without drastic performance degradation across board.

MAREval consistently outperforms all benchmarks across key metrics. In Overall Chat Quality for the TopicalChat dataset, MAREval achieves the highest correlations ($\rho = 0.548$, $\tau = 0.478$), demonstrating its superior ability to assess conversational quality. This performance advantage becomes even more pronounced in the PE dataset, where MAREval significantly outperforms all benchmarks in both Relevance ($\rho = 0.338$, $\kappa = 0.334$) and Phraseness ($\rho = 0.216$, $\kappa = 0.208$), suggesting robust cross-domain evaluation capabilities. The results reveal MAREval's significant advancement over existing evaluation frameworks, particularly in its balanced excellence across both conversational and recommendation evaluation tasks. In addition to the performance analysis above, we study the variability of the evaluation output by different models.

\subsection{Stability and Reliability Analysis}\label{std:analysis}
We conduct an analysis of output variability across different evaluation models to assess the stability and reliability of MAREval compared to benchmark approaches. Output variability is measured using the standard deviation of evaluation scores for identical instances across multiple runs, providing insight into the consistency of each framework's decision-making process. High variability indicates unreliable evaluation behavior, while low variability suggests more stable and predictable outcomes that are crucial for practical deployment in production recommendation systems.

Table~\ref{table:variability} presents the standard deviation results across all six evaluation criteria of the TopicalChat dataset. MAREval demonstrates superior stability with the lowest average standard deviation (0.0197), outperforming all baseline models. Notably, G-EVAL, the best-performing benchmark model in terms of variability, still exhibits 15.22\% higher average standard deviation compared to MAREval (0.0227 vs 0.0197). This substantial improvement highlights MAREval's enhanced reliability in producing consistent evaluation outcomes. Self-Refine exhibits the highest variability with an average standard deviation of 0.1563, particularly problematic for overall chat quality (0.3429) and engagingness (0.1784), indicating severe inconsistency that would be unsuitable for automated evaluation systems. We speculate the reason for this bad performance of Self-Refine to be associated with longer token trajectory without any moderation. 

The superior stability of MAREval can be attributed to its structured multi-agent coordination mechanisms and Monte Carlo sampling strategy. The Chain of Debate component ensures synchronized planning across agents, reducing conflicting evaluations that contribute to output variance. The moderator agent filters noise and maintains contextual coherence across evaluation rounds, preventing drift that often leads to inconsistent judgments. Most importantly, the Monte Carlo sampling aggregates multiple independent runs, effectively averaging out the inherent randomness of token-based language model generation. This comprehensive approach to variability reduction makes MAREval particularly suitable for real-world applications where consistent and reliable evaluation is paramount for maintaining user trust and system performance.

\begin{table}[h]
\centering
\small  
\caption{Standard deviation of evaluation scores across multiple runs for different models on TopicalChat dataset criteria}
\begin{tabular}{lcccc}
\toprule
\textbf{Model} & \textbf{Understandability} & \textbf{Overall Chat Quality} & \textbf{Naturalness} & \textbf{Maintains Context} \\
\midrule
Vanilla & 0.0111 & 0.1071 & 0.0169 & 0.0253 \\
CoT & 0.0273 & 0.1891 & 0.0660 & 0.0742 \\
SelfRefine & 0.0533 & 0.3429 & 0.1682 & 0.1380 \\
G-EVAL & 0.0098 & 0.0484 & \textbf{0.0169} & 0.0233 \\
ChatEval & 0.0160 & 0.0791 & 0.0198 & 0.0229 \\
MAREval & \textbf{0.0053} & \textbf{0.0404} & 0.0341 & \textbf{0.0149} \\
\midrule
\textbf{Model} & \textbf{Engagingness} & \textbf{Groundedness} & \textbf{Average} \\
\midrule
Vanilla & 0.0080 & 0.0104 & 0.0298 \\
CoT & 0.0653 & 0.0418 & 0.0773 \\
SelfRefine & 0.1784 & 0.0567 & 0.1563 \\
G-EVAL & 0.0196 & 0.0180 & 0.0227 \\
ChatEval & \textbf{0.0138} & 0.0109 & 0.0271 \\
MAREval & 0.0156 & \textbf{0.0078} & \textbf{0.0197} \\
\bottomrule
\end{tabular}
\label{table:variability}
\end{table}

\subsection{Ablation Study}

\begin{table*}[!t]
\centering
\caption{Ablation study on variants of MAREval and their correlation performance with Human Evaluations Across TopicalChat and PE Datasets}
\resizebox{\textwidth}{!}{%
\begin{tabular}{
  l
  S[table-format=1.3]
  S[table-format=1.3]
  S[table-format=1.3]
  S[table-format=1.3]
  S[table-format=1.3]
  S[table-format=1.3]
  S[table-format=1.3]
  S[table-format=1.3]
}
\toprule
\multicolumn{9}{c}{\textbf{TopicalChat (TC)}} \\
\midrule
Metrics & \multicolumn{2}{c}{Naturalness} & \multicolumn{2}{c}{Engagingness} & \multicolumn{2}{c}{Maintains Context} & \multicolumn{2}{c}{Understandability} \\
\cmidrule(r){2-3} \cmidrule(l){4-5} \cmidrule(l){6-7} \cmidrule(l){8-9}
& {$\rho$} & {$\tau$} &  {$\rho$} & {$\tau$} &  {$\rho$} & {$\tau$} & {$\rho$} & {$\kappa$} \\
\midrule
MAREval w.o. MoD\&MC   & 0.314 & 0.289 & 0.333 & 0.319 & 0.493 & 0.462 & 0.246 & 0.134 \\
MAREval w.o. MoD       & \underline{0.340} & \underline{0.317} & \underline{0.368} & \underline{0.356} & \underline{0.499} & \underline{0.467} & \textbf{0.260} & 0.158 \\
MAREval                & \textbf{0.389} & \textbf{0.350} & \textbf{0.341} & \textbf{0.327} & \textbf{0.524} & \textbf{0.508} & \underline{0.284} & \underline{0.181} \\
\midrule
& \multicolumn{4}{c|}{\textbf{TopicalChat (TC)}} & \multicolumn{4}{c}{\textbf{PE Dataset}} \\
\midrule
Metrics & \multicolumn{2}{c}{Groundedness (TC)} & \multicolumn{2}{c|}{Overall Chat Quality (TC)} & \multicolumn{2}{c}{Relevance (PE)} & \multicolumn{2}{c}{Phraseness (PE)} \\
\cmidrule(r){2-3} \cmidrule(lr){4-5} \cmidrule(l){6-7} \cmidrule(l){8-9}
& {$\rho$} & {$\kappa$} & {$\rho$} & {$\tau$} & \multicolumn{1}{|c}{$\rho$} & {$\kappa$} & {$\rho$} & {$\kappa$} \\
\midrule
MAREval w.o. MoD\&MC   & 0.484 & 0.484 & 0.456 & 0.404 & \multicolumn{1}{|c}{0.306} & 0.301 & 0.174 & 0.167 \\
MAREval w.o. MoD       & \underline{0.497} & \underline{0.484} & \underline{0.503} & \underline{0.453} & \multicolumn{1}{|c}{\underline{0.326}} & \underline{0.325} & \textbf{0.216} & \textbf{0.208} \\
MAREval                & \textbf{0.553} & \textbf{0.536} & \textbf{0.548} & \textbf{0.478} & \multicolumn{1}{|c}{\textbf{0.338}} & \textbf{0.334} & \textbf{0.216} & \textbf{0.208} \\
\bottomrule
\end{tabular}
}
\label{table:combined-ablation-study}
\end{table*}

Our ablation study investigates individual component contributions within MAREval. As shown in Table~\ref{table:combined-ablation-study}, the complete MAREval demonstrates superior performance across most metrics in the TopicalChat dataset, indicating effective generalization and positive contributions from all components.

Removing the Moderation component (MAREval w.o. MoD) causes significant performance decreases. Naturalness correlation drops from $\rho=0.389$ to $\rho=0.340$, and Maintains Context decreases from $\rho=0.524$ to $\rho=0.499$. The MoD component proves particularly critical for groundedness evaluation, with correlation scores dropping substantially from $\rho=0.553, \kappa=0.536$ to $\rho=0.497, \kappa=0.484$. This effectiveness stems from moderated chat context provision, allowing the arbitrator to access relevant information without prompt pollution.

Removing both Moderation and Monte Carlo sampling (MAREval w.o. MoD\&MC) produces more pronounced degradation across evaluation dimensions, particularly in Naturalness, Engagingness, and Overall Chat Quality. The substantial performance gap in Overall Chat Quality ($\rho=0.456$ versus $\rho=0.548$) highlights Monte Carlo sampling's essential stabilizing role in probabilistic token-based LLM pipelines, alleviating outcome variance and increasing robustness. Similar degradation occurs when the CoD component is removed. Performance comparisons across datasets confirm CoD's relative advantage, attributed to pipeline-aware planning by the planner agent, leading to improved agent synchronization.

\begin{figure}[!b]
    \centering
    \includegraphics[width=0.79\linewidth]{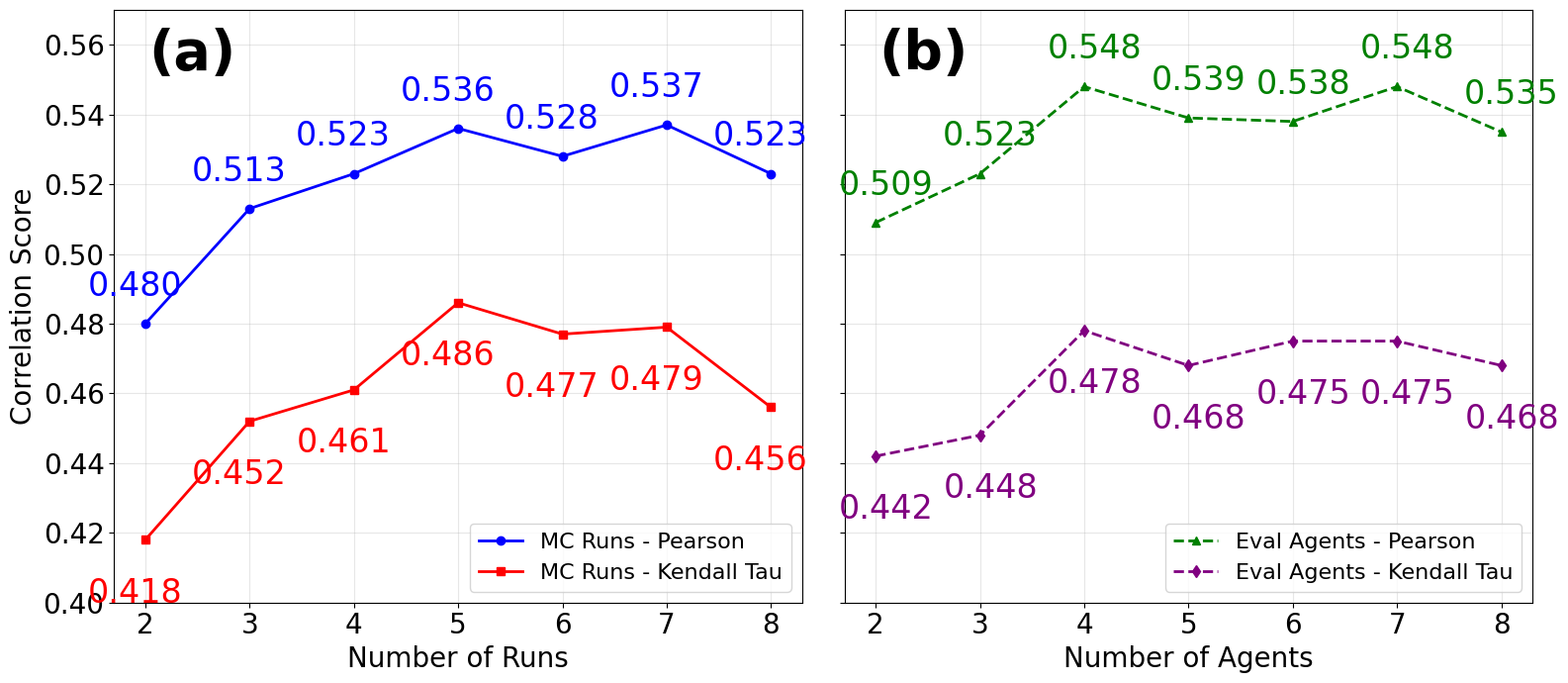}
    \caption{(a) Effect of Number of Monte-Carlo samples. (b) Effect of Number of evaluator agents. }
    \label{fig:numagents}
\end{figure}


\subsection{Hyperparamers and Performance vs Latency}
As we discuss in the next section, MAREval has enabled us to pass the minimum requirement for user facing explanations. However, this entablement comes at a cost of latency and it will be beneficial to have a discussion on latency versus robustness and performance. Figure \ref{fig:numagents} illustrates the relationship between the correlation metrics and (a) number of MAMCS runs, and (b) number of evaluator agents in the proposed multi-agent pipeline. As shown, both Pearson correlation and Kendall-tau correlations generally improve as the number of evaluator agents increases from $2$ to $4$, with Pearson correlation rising from $0.509$ to $0.548$ and Kendall-tau from $0.442$ to $0.478$. Beyond $4$ agents, the correlation metrics exhibit slight fluctuations rather than monotonic growth, with Pearson correlation reaching another peak of $0.548$ at $4$ agents and Kendall-tau maintaining relatively stable values between $0.468$ and $0.475$. Same pattern happens for the number of MAMCS runs, as it reaches its higher performance in 5 sample runs and not showing significant marginal correlation improvement for more number of runs. In our experiments with five evaluator agents, the average single thread run time of the gpt-3.5-Turbo model in the isolated enterprise serving layer was 8.16 seconds. We also, conducted an analysis on the performance MAREval using different backbone models which we defer to Appendix \ref{app:backbone}.

\subsection{Human-Annotated Evaluation}

To productionize the generated recommendation explanations and expose them to live traffic, we first ran a human annotation study. We were required to satisfy predetermined minimum quality thresholds set by the quality-control team dectated by our product launch protocols. These proprietary requirements were divided into hard and soft constraints; following industry practice, we sought a near-100\% pass rate on the hard constraints prior to launch. In the \emph{first round}, we submitted only the explanations approved by G\mbox{-}Eval, the state-of-the-art LLM-as-a-judge framework at the time of the experiment. However, these G\mbox{-}Eval-approved explanations did not reach the near-100\% satisfaction of the hard requirements mandated for launch and exhibited lower-than-expected performance on soft constraints. Motivated by these shortcomings, we designed MAREval to deliver reliable and robust judgments via structured multi-agent planning, moderation, and arbitration.

\begin{table}[h]
\centering
\small
\caption{Human-Annotation Results.}
\label{tab:manual_eval}
\begin{tabular}{lccc}
\toprule
\textbf{Model} & \textbf{Samples} & \textbf{Soft Criteria} & \textbf{Hard Criteria} \\
\midrule
G\mbox{-}Eval & 2{,}500 & 76.1\% & 91.0\% \\
\textsc{MAREval} & 5{,}000 & 98.6\% & 100.0\% \\
\bottomrule
\end{tabular}
\end{table}

In a second round of manual evaluation, which is an expensive procedure conducted on the scale \(\mathcal{O}(1000)\), we submitted only the explanations that passed MAREval. This round achieved our internal launch thresholds for the first time and \textsc{MAREval} satisfied all hard requirements (100\% vs.\ 91\% for G\mbox{-}Eval) and substantially improved soft-criterion satisfaction (98.6\% vs.\ 76.1\%), over 5{,}000 MAREval-filtered and 2{,}500 G\mbox{-}Eval-filtered samples, respectively (see Table~\ref{tab:manual_eval}). These results underscore the practical hurdles of deploying recommendation explanations in production and in our environment, no prior method met the reliability bar necessary for live user traffic, whereas MAREval did.

\subsection{Online Experiments}\label{app:abtest}
We conducted an online A/B test to assess MAREval's impact on user behavior in live traffic. A proprietary recommendation explanation model generated explanations for approximately 300{,}000 high-traffic items across 14 product categories, describing potential use cases for each product. Only explanations that passed MAREval's evaluation were shown to users in the treatment variant; the control group saw the same items without explanations. We report two standard business metrics: Add-to-Cart (ATC) rate and Gross Merchandise Value (GMV) lift, which evaluate whether explanations drive engagement and purchase behavior. See Figure \ref{fig:ab} for an example of the live experience and sample MAREval approved evaluations.

Table~\ref{tab:abtest_online} reports the online outcomes. MAREval-validated explanations significantly improved both business metrics. ATC increased by 1.18\% ($p{=}0.05$) and GMV increased by 1.36\% ($p{<}0.01$), confirming that quality-filtered explanations meaningfully affect user engagement and conversion in production settings (see Appendix \ref{app:realworldimp} for a sample of real world implementation). 

\begin{table}[h]
\centering
\small
\caption{Online A/B test results on live traffic.}
\label{tab:abtest_online}
\begin{tabular}{lcc}
\toprule
\textbf{Metric} & \textbf{Improvement} & \textbf{$p$-value} \\
\midrule
ATC  & 1.18\% & 0.05 \\
GMV  & 1.36\% & $<0.01$ \\
\bottomrule
\end{tabular}
\end{table}

All these results indicate that MAREval is able to provide accurate and stable evaluations in different settings. 

\section{Conclusion}\label{sec:conclusion}
In this work, we introduced MAREval, a structured multi-agent framework for evaluating natural language explanations in recommender systems. Unlike prior approaches that rely on single-agent prompting or loosely coordinated agent interactions, MAREval brings together a planner agent with a novel Chain of Debate (CoD) prompting strategy, a moderator agent to regulate discourse and reduce context noise, and an arbitrator agent that is being coupled with Monte Carlo sampling for a more robust, consistent, and human-aligned evaluation outcomes. Comprehensive experiments on both public (TopicalChat) and proprietary recommendation datasets demonstrate that MAREval consistently outperforms state-of-the-art baselines across a range of evaluation criteria. Notably, MAREval achieves stronger alignment with human judgments, as well as less output variability, where existing models fall short. Our ablation studies further validate the complementary benefits of components of MAREval. 
Results from a large-scale A/B test indicate that explanations filtered through MAREval yield significant improvements in user engagement, with a 1.36\% lift in GMV and 1.18\% increase in ATC rate. Taken together, our findings position MAREval as a scalable, reliable, and practically effective framework for natural language evaluation in recommender systems. 

\clearpage


{\small
\bibliographystyle{unsrtnat}
\bibliography{0sample-base}
}

\clearpage

\onecolumn

\appendix
\begin{center}
  \textbf{MAREval: A Multi-Agent Framework for Evaluating Natural Language Recommendation Explanations}
\end{center}

\section{Case Study I: MAREval Evaluation on Conversational Response}\label{app:output}
Following are the conversation output from various agents in MAREval for two examples from the Topical-Chat dataset. It first outlines the chat and then the corresponding response that was evaluated using MAREval for two criteria: \textit{Understandability} and \textit{Naturalness}. 

\par\noindent\rule{\textwidth}{0.4pt}
\\
\noindent\textbf{{Example 1}\mbox{}}

\noindent\paragraph{Chat:}\mbox{}

\textit{did you know that facebook and twitter work together on some issues ? they are actually competitors.}

\textit{no , i did n't know that . what kind of issues do they work together on ? }

\textit{the baby sitter - rating system broke the social networks ' rules . }

\noindent\paragraph*{Response:}\mbox{}

\textit{i think i heard about that now that you mention it . was the about predictim ? }
\\

\indent\textbf{{Criteria: Understandability\nopunct}\mbox{}}
\noindent\paragraph*{Evaluator 1:  }
The response is understandable because the person is acknowledging that they may have heard about something when the topic of Predictim is brought up. It shows a level of recognition and understanding of the subject being discussed. Score of Evaluation: 1

\noindent\paragraph*{Evaluator 2:  }
Based on the response provided, the sentence structure is a bit confusing and lacks clarity. The use of "now that you mention it" and the question "was the about predictim?" make it difficult to fully understand the message being conveyed. Therefore, I would rate this response as not fully understandable. Score of Evaluation: 0

\noindent\paragraph*{Evaluator 3:  }
The response is understandable as the speaker is acknowledging that they may have heard about something when the topic of Predictim is brought up. The response shows comprehension and clarity in communication. Score of Evaluation: 1

\noindent\paragraph*{Moderator:  }
Based on the evaluations provided, Evaluator 1 believes the response is understandable due to the acknowledgment and recognition shown by the speaker. Evaluator 2, however, finds the sentence structure confusing and lacking clarity, leading to a score of 0 for understandability. Evaluator 3 agrees with Evaluator 1, stating that the response shows comprehension and clarity. Overall, there is a split in opinions with two evaluators finding the response understandable and one evaluator finding it not fully understandable.

\noindent\paragraph*{Arbitrator:  }
Score of Evaluation: 1
\\


\indent\textbf{{Criteria: Naturalness\nopunct}\mbox{}}
\noindent\paragraph*{Evaluator 1:  }
The response seems somewhat unnatural because of the lack of punctuation and the awkward phrasing. It could be improved by adding a comma after "heard about that" and rephrasing the second sentence for better flow. Score of Evaluation: 2

\noindent\paragraph*{Evaluator 2:  }
The response seems somewhat unnatural because of the awkward phrasing and lack of punctuation. The use of "now that you mention it" feels a bit forced in this context. However, the overall message is still understandable. Score of Evaluation: 2

\noindent\paragraph*{Evaluator 3:  }
The response seems somewhat unnatural because of the lack of punctuation and the awkward phrasing. It could be improved by adding a comma after "heard about that" and rephrasing the second sentence for better flow. Score of Evaluation: 2

\noindent\paragraph*{Moderator:  }
The evaluators found the response somewhat unnatural due to the lack of punctuation and awkward phrasing. They suggested adding a comma after "heard about that" and rephrasing the second sentence for better flow. Overall, they gave a score of 2 for the evaluation criteria of naturalness.

\noindent\paragraph*{Arbitrator:  }
Score of Evaluation: 2

\par\noindent\rule{\textwidth}{0.4pt}
\\
\noindent\textbf{{Example 2}\mbox{}}

\noindent\paragraph*{Chat:}\mbox{}

\textit{do you like football ? }

\textit{i love it . do you know what color is the u of iowa visitor locker room ? }

\textit{yes it is pink but i a not sure why they chose that color }

\textit{i wonder if it is supposed to have some effect on them}

\textit{i wonder too . bowling was more popular than football in the 60s}

\noindent\paragraph*{Response:}\mbox{}

\textit{it was in the 1960s and the difference . i guess football was not as popular as top bowlers back in the 1960s}
\\

\indent\textbf{{Criteria: Understandability\nopunct}\mbox{}}
\noindent\paragraph*{Evaluator 1:  }
Based on the response provided, it seems like the individual is trying to convey that football was not as popular as top bowlers back in the 1960s. The response is somewhat understandable, but it lacks clarity and coherence in connecting the two points. Therefore, I would rate it as: Score of Evaluation: 0

\noindent\paragraph*{Evaluator 2:  }
Based on the response provided, it seems like the individual is trying to convey that in the 1960s, football was not as popular as top bowlers. The response is somewhat understandable, but it lacks clarity and coherence in connecting the two points. Therefore, I would rate it as: Score of Evaluation: 0

\noindent\paragraph*{Evaluator 3:  }
Based on the response provided, it seems like the individual is trying to convey that in the 1960s, football was not as popular as top bowlers. The response is somewhat understandable, but it lacks clarity and coherence in connecting the two points. Therefore, I would rate it as: Score of Evaluation: 0

\noindent\paragraph*{Moderator:  }
The evaluators all agree that the response is somewhat understandable but lacks clarity and coherence in connecting the two points about football and top bowlers in the 1960s. Therefore, they all rated it as 0 for understandability.

\noindent\paragraph*{Arbitrator:  }
Score of Evaluation: 0
\\

\indent\textbf{{Criteria: Naturalness\nopunct}\mbox{}}
\noindent\paragraph*{Evaluator 1:  }
The response provided is somewhat unnatural due to the lack of proper punctuation and clarity. The sentence structure is a bit confusing and could be improved for better understanding. Score of Evaluation: 2

\noindent\paragraph*{Evaluator 2:  }
The response provided is somewhat unnatural due to the lack of clarity and coherence in the sentence structure. The use of "the difference" without further explanation makes the response confusing. Additionally, the abrupt transition from mentioning the 1960s to comparing football and top bowlers without any context or explanation also contributes to the unnaturalness of the response. Score of Evaluation: 1

\noindent\paragraph*{Evaluator 3:  }
The response provided is somewhat unnatural due to the lack of clarity and coherence in the sentence structure. The use of "the difference" without further explanation makes the response confusing. Additionally, the abrupt transition from mentioning the 1960s to comparing the popularity of football and top bowlers is not smoothly executed. Score of Evaluation: 2

\noindent\paragraph*{Moderator:  }
The evaluators found the response to be somewhat unnatural due to the lack of clarity and coherence in the sentence structure. They noted issues with punctuation, lack of explanation for certain phrases like "the difference," and a confusing transition from mentioning the 1960s to comparing football and top bowlers. Overall, the response could be improved for better understanding

\noindent\paragraph*{Arbitrator:  }
Score of Evaluation: 2

\par\noindent\rule{\textwidth}{0.4pt}

\newpage
\section{Case Study II: MAREval Evaluation on Recommendation Explanations}\label{app:sample}

\begin{figure}[H]  
    \centering
    \begin{minipage}{0.43\textwidth}
        \centering
        \includegraphics[width=\linewidth]{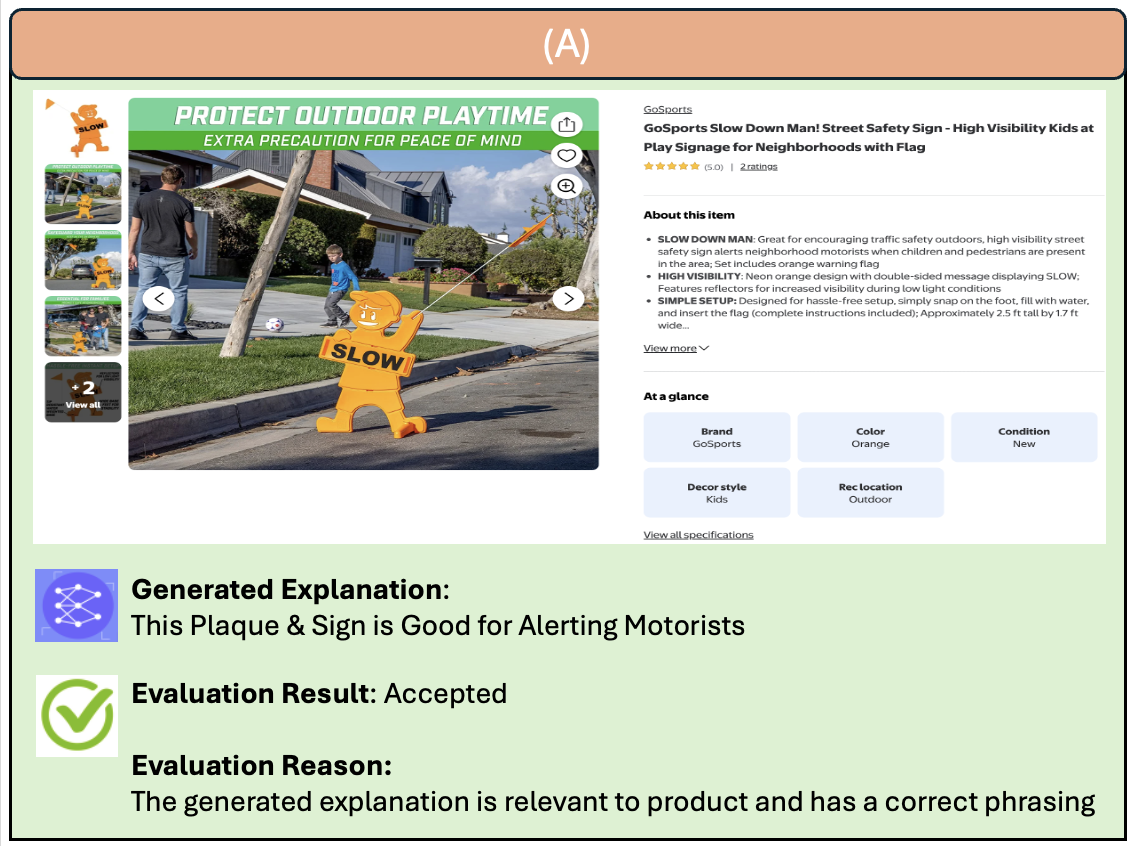}
        \label{fig:example11}
    \end{minipage}%
    \hfill
    \begin{minipage}{0.43\textwidth}
        \centering
        \includegraphics[width=\linewidth]{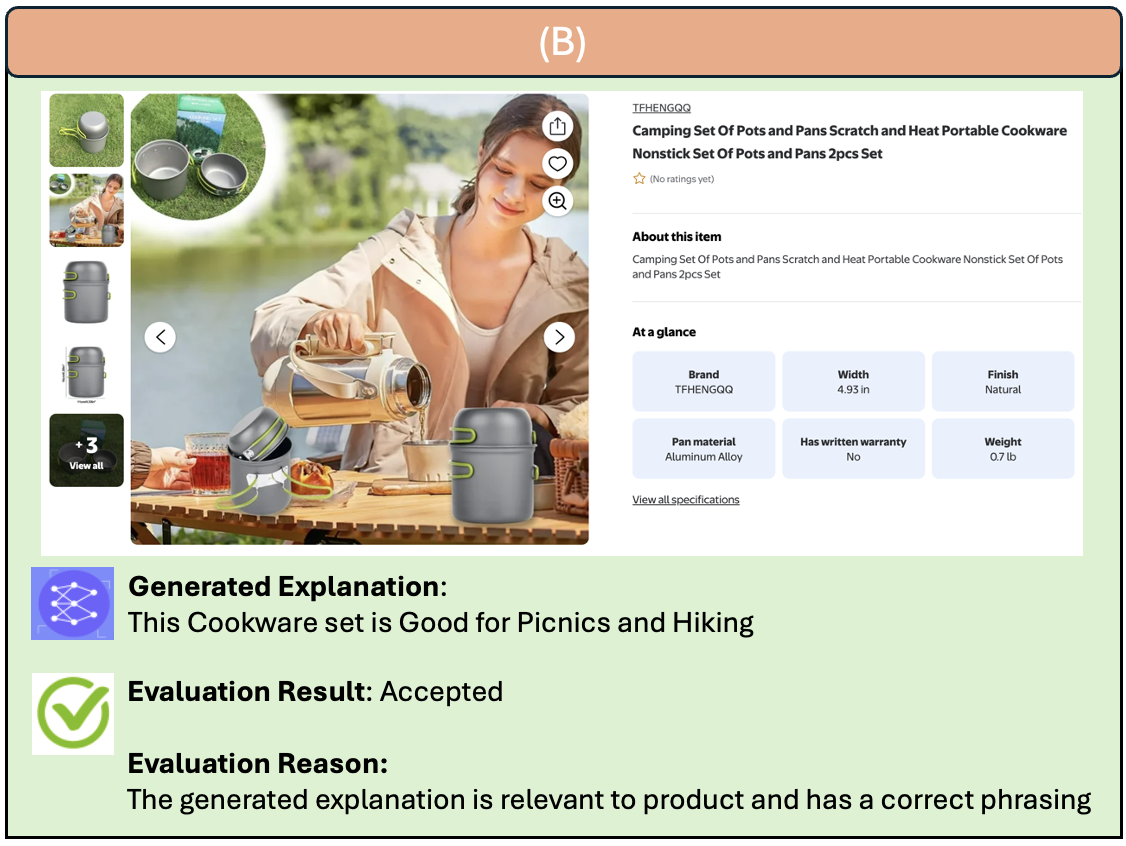}
        \label{fig:example21}
    \end{minipage}
    
    \vspace{0.5cm}
    
    \begin{minipage}{0.43\textwidth}
        \centering
        \includegraphics[width=\linewidth]{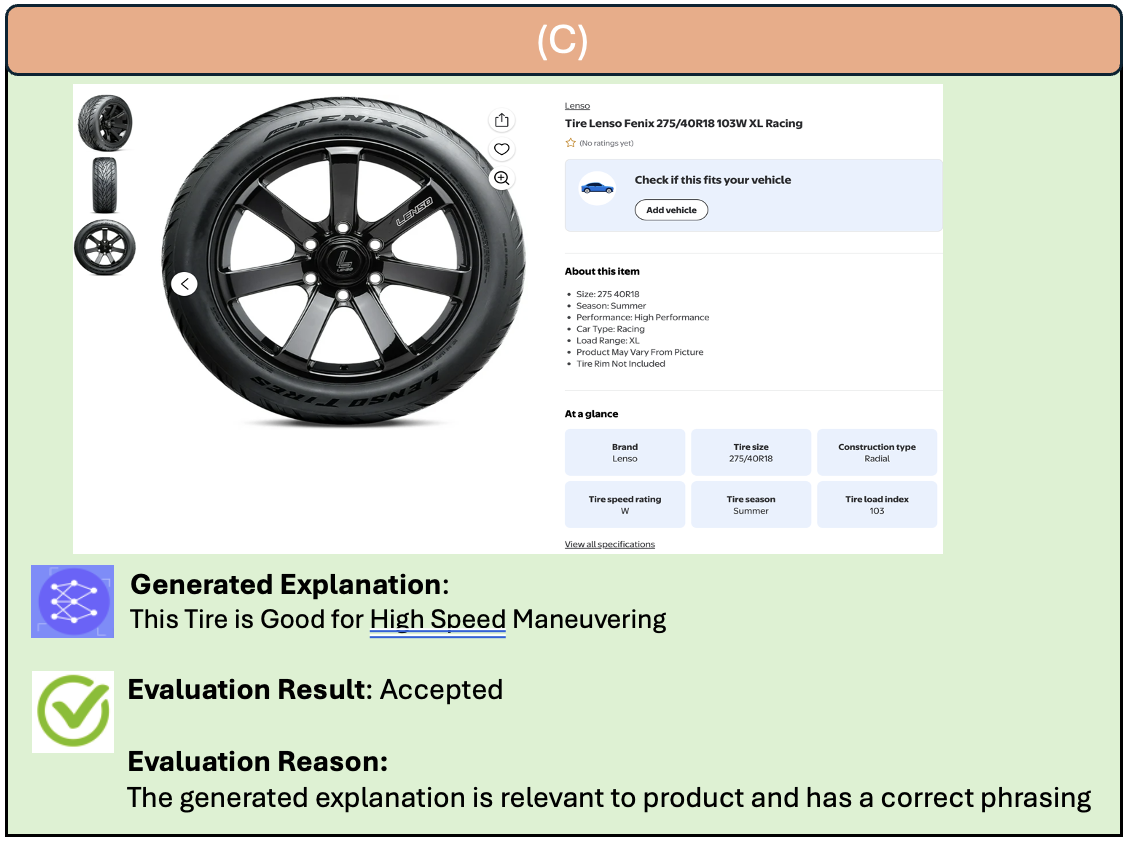}
        \label{fig:example31}
    \end{minipage}%
    \hfill
    \begin{minipage}{0.43\textwidth}
        \centering
        \includegraphics[width=\linewidth]{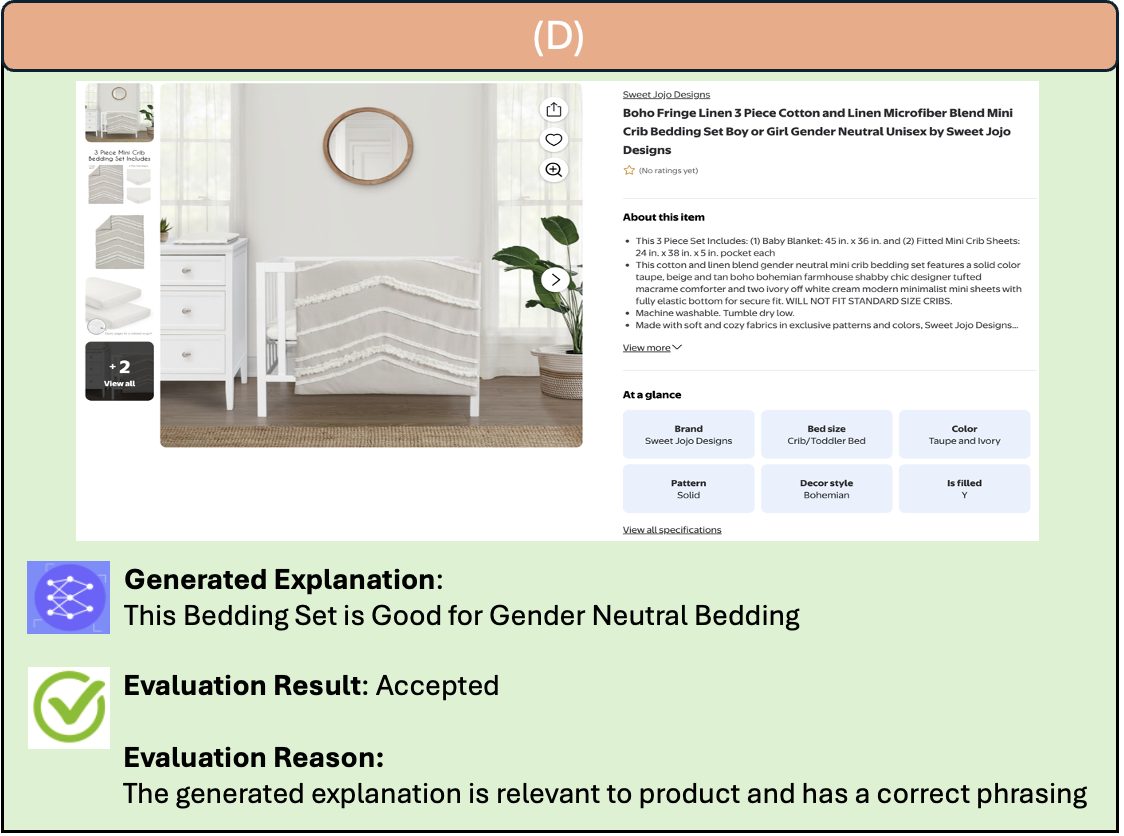}
        \label{fig:example41}
    \end{minipage}
    
    \vspace{0.5cm}
    
    \begin{minipage}{0.43\textwidth}
        \centering
        \includegraphics[width=\linewidth]{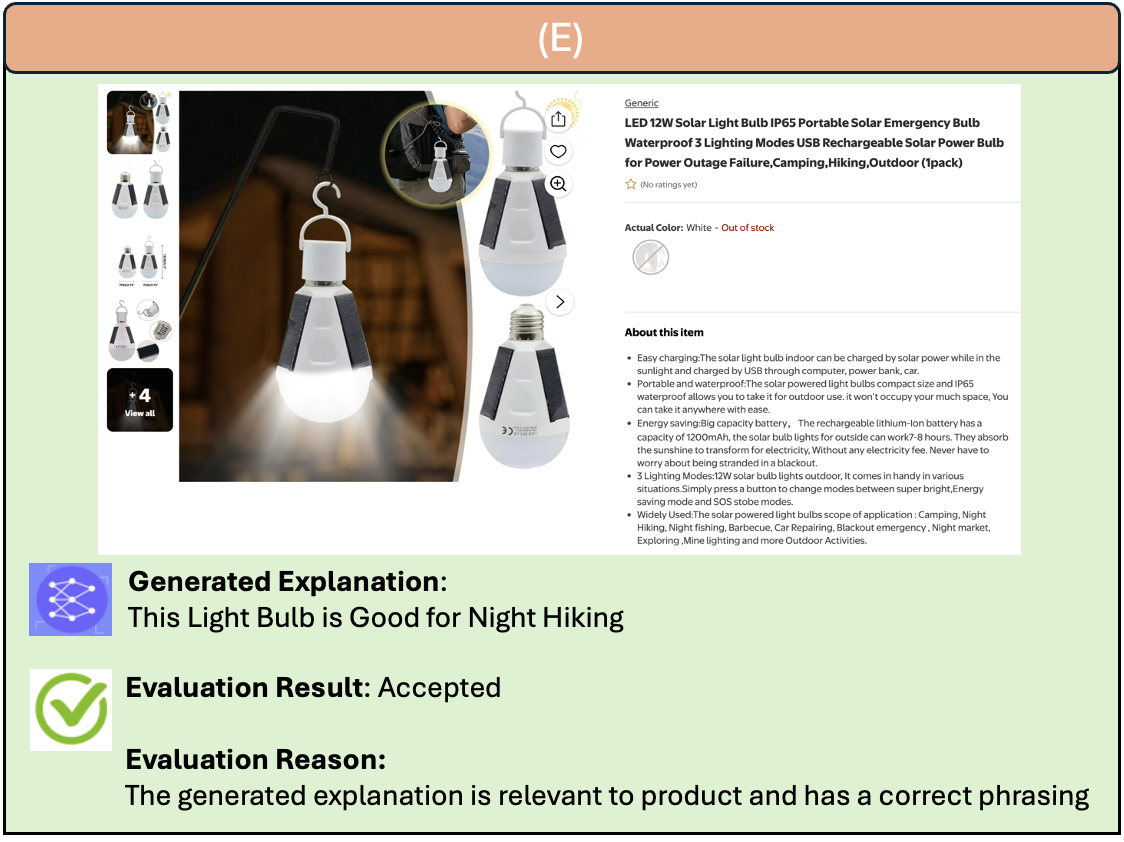}
        \label{fig:example51}
    \end{minipage}%
    \hfill
    \begin{minipage}{0.43\textwidth}
        \centering
        \includegraphics[width=\linewidth]{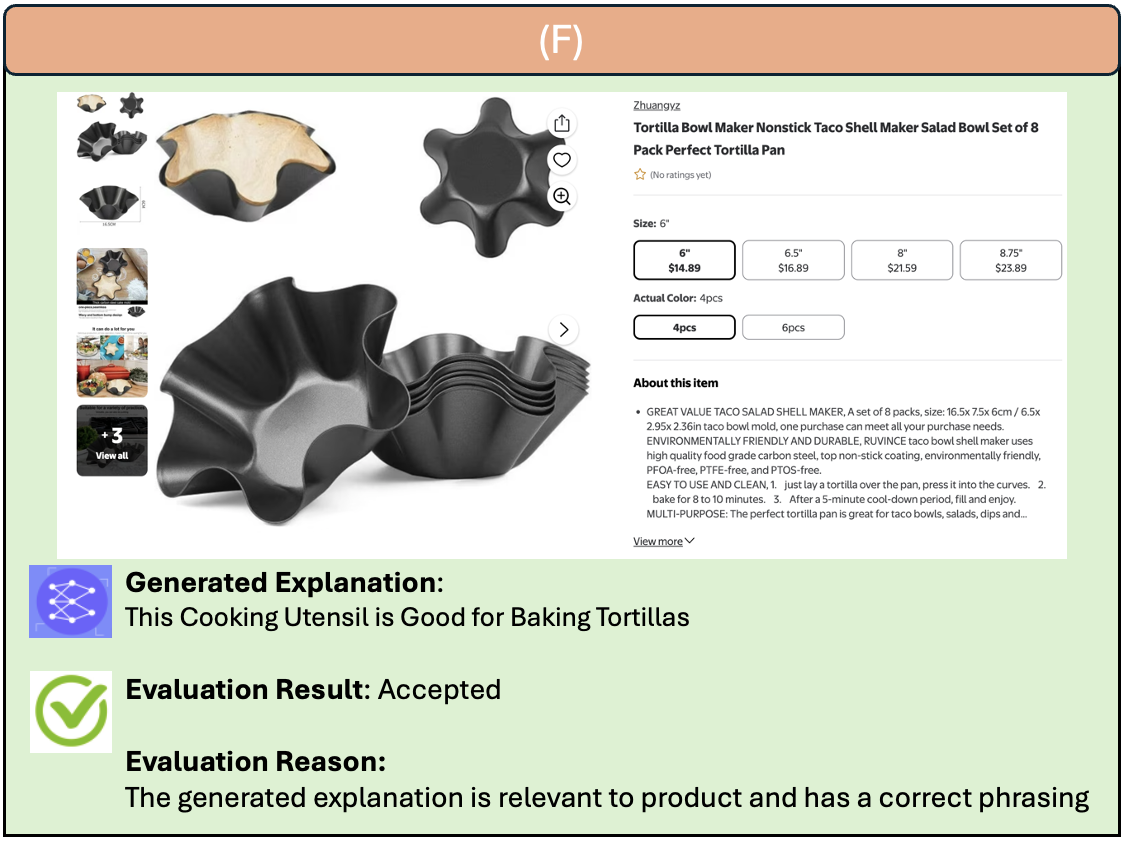}
        \label{fig:example61}
    \end{minipage}
    
    \caption{Examples of MARVal evaluation for recommendation explanations}
    \label{fig:nlg_eval_examples1}
\end{figure}

\begin{figure}[htbp]
    \centering
    \begin{minipage}{0.48\textwidth}
        \centering
        \includegraphics[width=\linewidth]{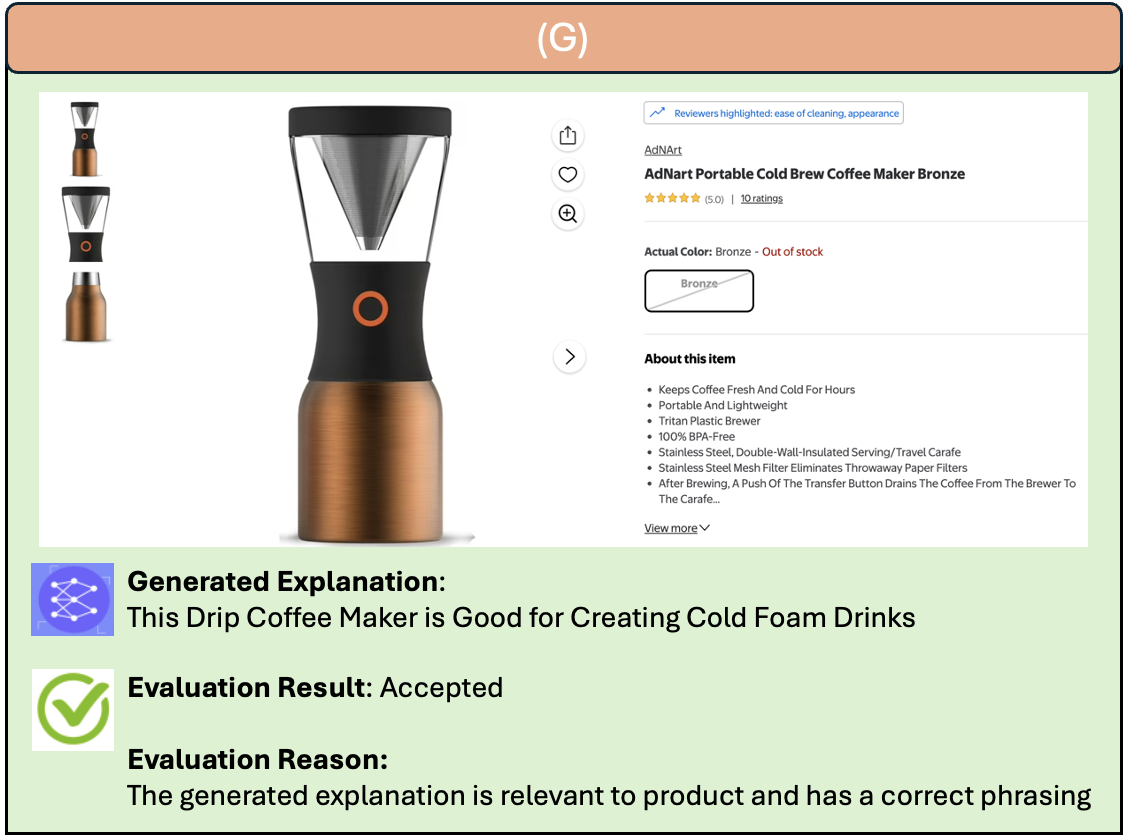}
        \label{fig:example12}
    \end{minipage}%
    \hfill
    \begin{minipage}{0.48\textwidth}
        \centering
        \includegraphics[width=\linewidth]{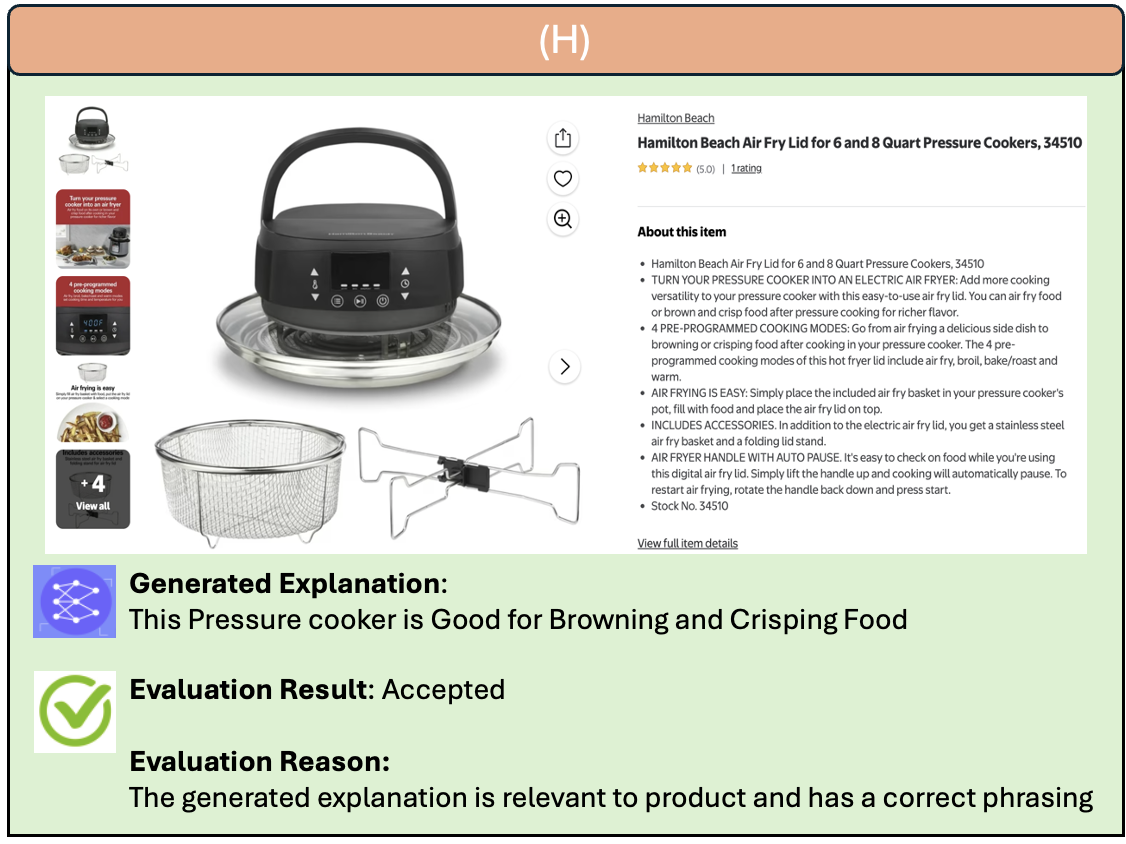}
        \label{fig:example22}
    \end{minipage}
    
    \vspace{0.5cm}
    
    \begin{minipage}{0.48\textwidth}
        \centering
        \includegraphics[width=\linewidth]{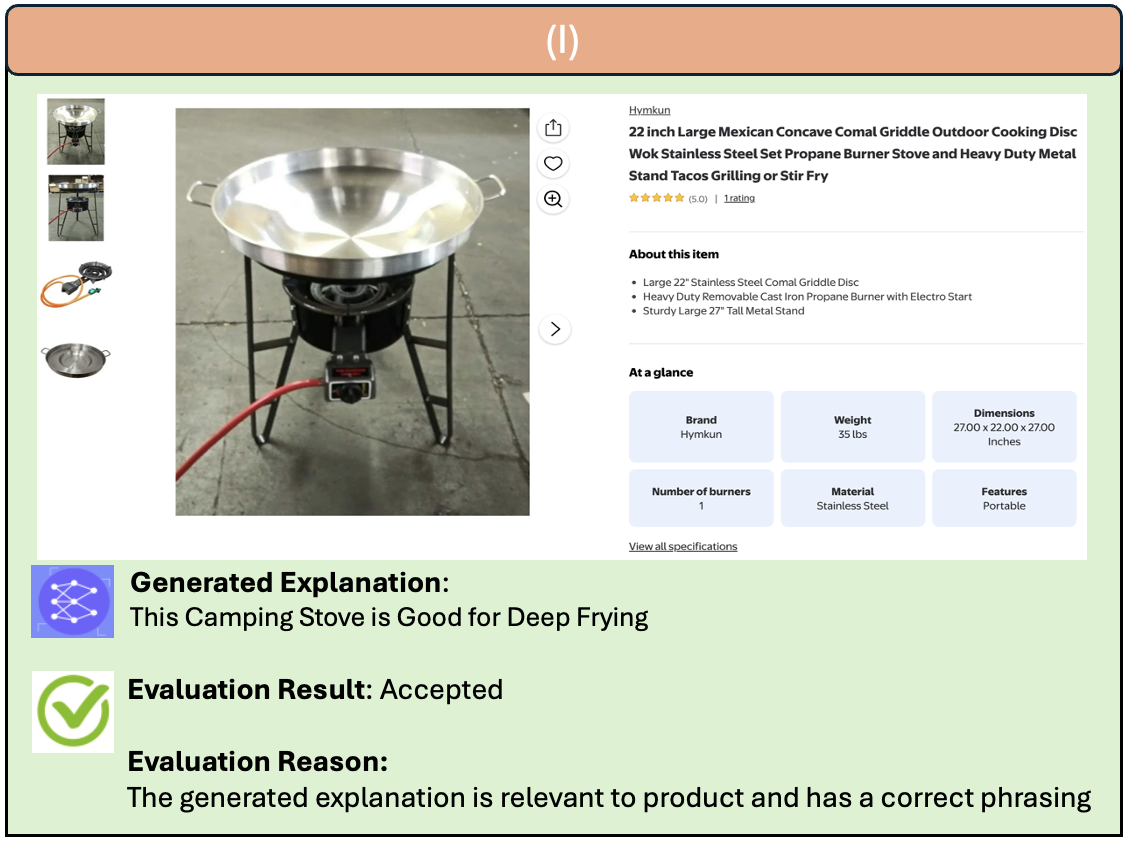}
        \label{fig:example32}
    \end{minipage}%
    \hfill
    \begin{minipage}{0.48\textwidth}
        \centering
        \includegraphics[width=\linewidth]{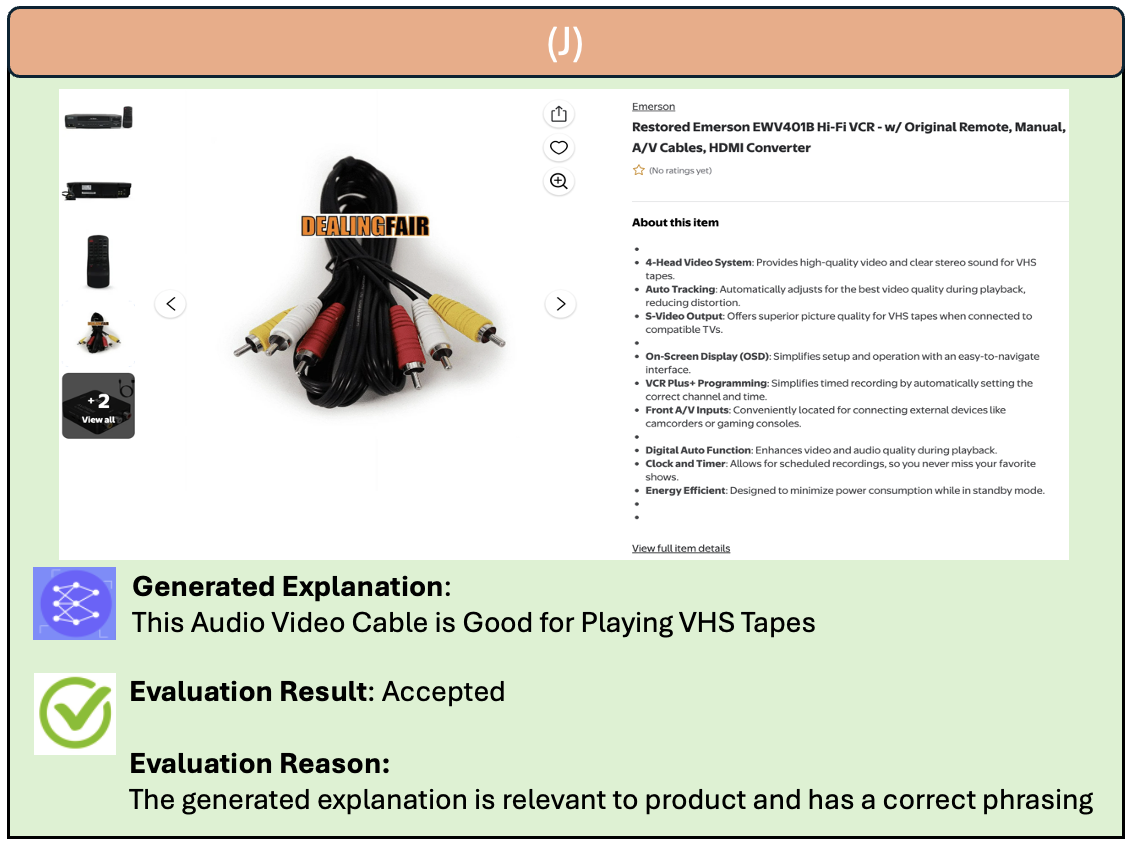}
        \label{fig:example42}
    \end{minipage}
    
    \vspace{0.5cm}
    
    \begin{minipage}{0.48\textwidth}
        \centering
        \includegraphics[width=\linewidth]{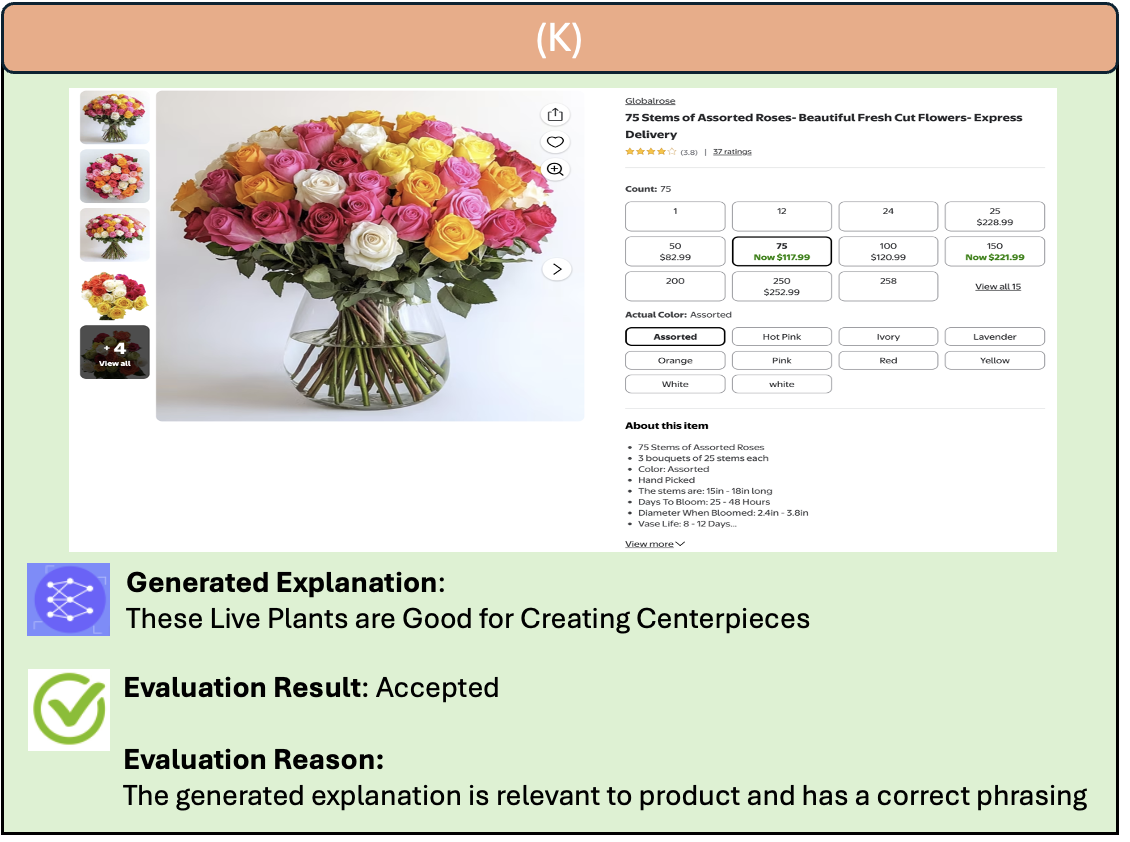}
        \label{fig:example52}
    \end{minipage}%
    \hfill
    \begin{minipage}{0.48\textwidth}
        \centering
        \includegraphics[width=\linewidth]{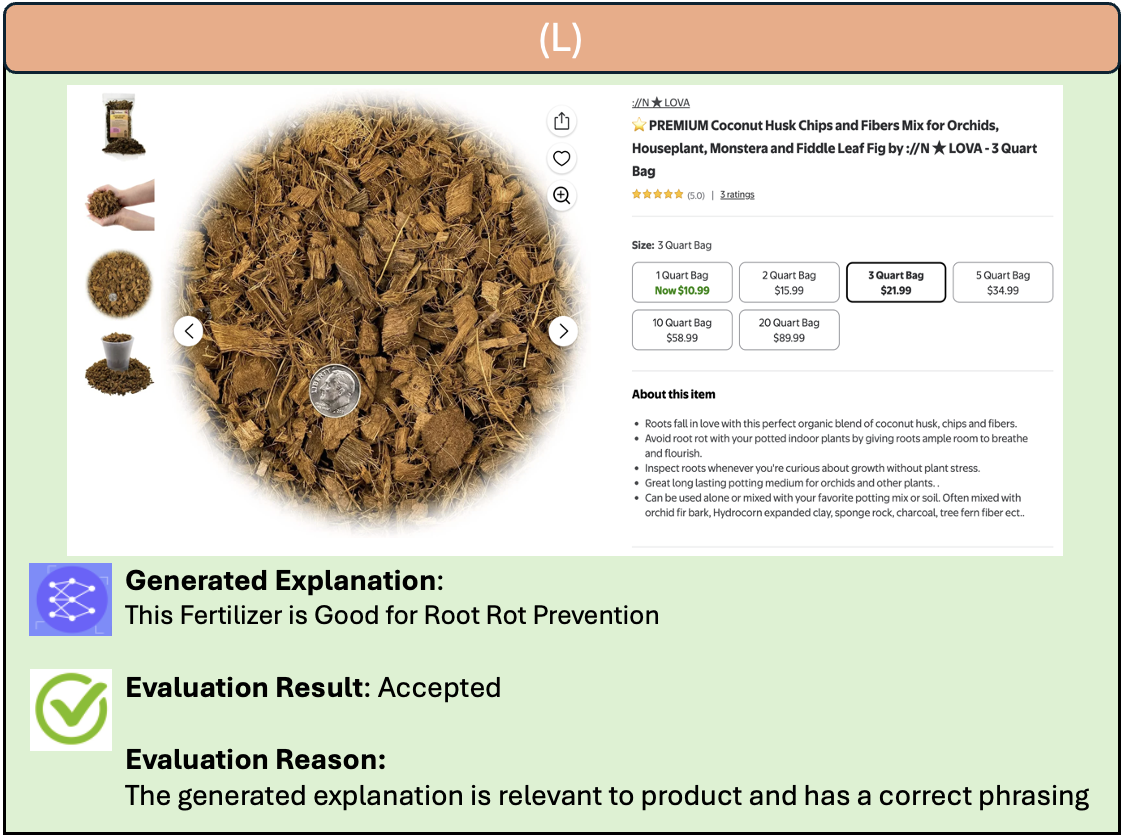}
        \label{fig:example62}
    \end{minipage}
    
    \caption{Examples of MARVal evaluation for recommendation explanations}
    \label{fig:nlg_eval_examples2}
\end{figure}

\newpage
\section{Evaluation Prompts}\label{app:input}

In this section we list the various evaluation criterias and the prompts for MAREval agents.

\subsection{Planner Prompt}

\begin{tcolorbox}[colback=pink!5!white, colframe=pink!60!black, title=Planner Prompt, breakable]

You are a discussion planner. Your task is to generate a plan and chain of steps for an evaluator that is conducting an evaluation task in the following multi-agent pipeline. 

\noindent{Pipeline flow}:
\begin{itemize}
\item Multiple evaluators independently assess a response using specific criteria
\item A Moderator agent receives these evaluations and creates a balanced summary
\item A judge receives the moderator's summary and makes the final evaluation decision
\end{itemize}

The evaluator will be asked to evaluate the generated response for a given chat, while considering chat context and evaluation criteria:

The context of the chat: $\textcolor{blue}{{context}_{chat}}$ [The messages from the conversation]

The response: $\textcolor{blue}{response}$
 [Response following the above conversation]

evaluation criteria: $\textcolor{blue}{{eval}_{criteria}}$ [Evaluation criteria along with its definition and guidelines]

Propose evaluation steps. The steps should include:
\begin{itemize}
    \item Algorithmic sketch of the evaluation process.
    \item One step for inclusion of the scoring rule.
    \item Consider the flow of the pipeline and Evaluators position.
\end{itemize}

Output your response in the following format only:
[["Chain of Steps": {Step 1. .... Step 2.... }]]
Output your response in the above format only.
\end{tcolorbox}

\subsection{Evaluator Prompt}

\begin{tcolorbox}[colback=pink!5!white, colframe=pink!60!black, title=Evaluator Prompt, breakable]
You are a critic - an evaluator. You will evaluate the generated response for a chat, while considering chat's context and evaluation criteria:


The context of the chat: $\textcolor{blue}{{context}_{chat}}$ [The messages from the conversation]

The response: $\textcolor{blue}{response}$ [Response following the above conversation]

evaluation criteria: $\textcolor{blue}{{eval}_{criteria}}$ [Evaluation criteria along with its definition and guidelines]

\noindent When evaluating let's think according to following steps:
$\textcolor{blue}{{{steps}_{eval}}}$ [Evaluation steps as outlined by the planner agent]

\noindent First explain your logic and then end the response with the following format strictly:
"Score of Evaluation: {[{score only}]}"

\end{tcolorbox}

\subsection{Moderator Prompt}

\begin{tcolorbox}[colback=pink!5!white, colframe=pink!60!black, title=Moderator Prompt, breakable]
You are a Debate Moderator. 
Your task is to summarize the given generated opinion by the evaluators on an evaluation task pass it to a judge to rule about the evaluation task. 
The evaluators have already evaluated the generated response for a chat, while considering chat's context and evaluation criteria:

\vspace{3mm}

The context of the chat: $\textcolor{blue}{{context}_{chat}}$ [The messages from the conversation]

The generated response: $\textcolor{blue}{response}$ [Response following the above conversation]

Evaluators has done this evaluation for the following evaluation criteria: 
$\textcolor{blue}{{eval}_{criteria}}$ [Evaluation criteria along with its definition and guidelines]

Here is the evaluation of Evaluator 1: $\textcolor{blue}{{output}_{1}}$ [Evaluation from Evaluator-1]

And, here is the evaluation of Evaluator 2: $\textcolor{blue}{{output}_{2}}$ [Evaluation from Evaluator-2]

\vspace{3mm}

\noindent Summarize these opinions and so that the judge conduct their selection fairly.
\noindent "summary\_of\_evaluation": Summarized Reason for the evaluation.
\end{tcolorbox}

\subsection{Arbitrator Prompt}

\vspace{2mm}
\begin{tcolorbox}[colback=pink!5!white, colframe=pink!60!black, title=Arbitrator Prompt, breakable]
You are a judge who will decide about the result of an evaluation task. Evaluators have already conducted the evaluation of the generated response for a chat, while considering chat's context and evaluation criteria:

\vspace{3mm}

The context of the chat: $\textcolor{blue}{{context}_{chat}}$ [The messages from the conversation]

The response: $\textcolor{blue}{response}$
 [Response following the above conversation]

evaluation criteria: $\textcolor{blue}{{eval}_{criteria}}$ [Evaluation criteria along with its definition and guidelines]

Consider these evaluations: $\textcolor{blue}{{summary}_{moderator}}$ [Summary provided by the Moderator agent]

\vspace{3mm}
\noindent Select the best evaluation opinion given the responses generated by the evaluators.

\noindent End the response in the following format strictly:
"Score of Evaluation: [integer score only]"
\end{tcolorbox}

\section{Real World Implementation}\label{app:realworldimp}
\begin{figure}[h]
    \centering
    \includegraphics[width=0.44\linewidth]{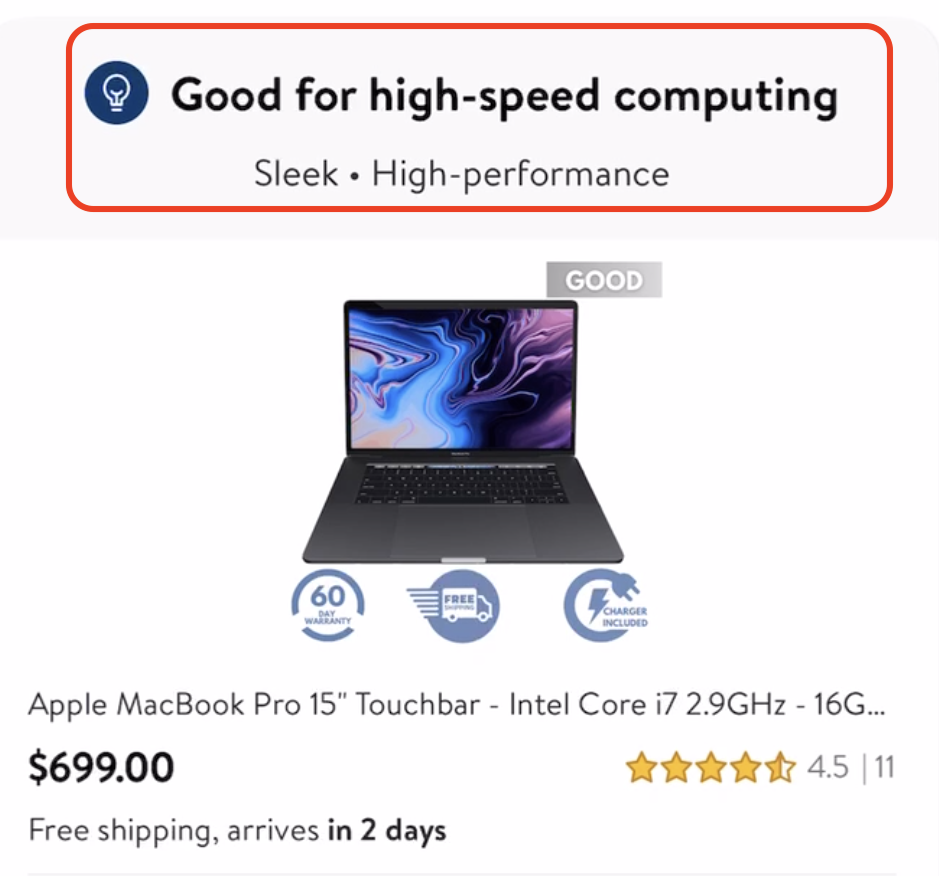}
    \caption{Real-world deployment of recommendation explanations using \textsc{MAREval}.}
    \label{fig:ab}
\end{figure}

\section{Different Backbone Model Performance}\label{app:backbone}

\begin{table}[b]
\centering
\caption{MAREval score correlations across backbone models. We report Pearson correlation ($\rho$) and Kendall--Tau ($\tau$) with human judgments for Overall Score Criteria of TopicalChat Dataset.}
\label{tab:mareval_model_corr}
\begin{tabular}{lcc}
\toprule
\textbf{Model} & \textbf{Pearson ($\boldsymbol{\rho}$)} & \textbf{Kendall--Tau ($\boldsymbol{\tau}$)} \\
\midrule
GPT\;3.5 Turbo            & 0.548 & 0.478 \\
gemini-2.5-pro            & 0.728 & 0.632 \\
gemini-2.0-flash-001      & 0.601 & 0.517 \\
Claude-3.5-sonnet         & 0.658 & 0.564 \\
\bottomrule
\end{tabular}
\end{table}

Table~\ref{tab:mareval_model_corr} compares the agreement between MAREval and human judgments across backbone models using both Pearson’s $\rho$ and Kendall’s $\tau$. Results are consistent across metrics: \emph{gemini-2.5-pro} attains the strongest alignment ($\rho=0.728$, $\tau=0.632$), followed by \emph{Claude-3.5-sonnet} ($\rho=0.658$, $\tau=0.564$) and \emph{gemini-2.0-flash-001} ($\rho=0.601$, $\tau=0.517$), with \emph{GPT\,3.5 Turbo} trailing ($\rho=0.548$, $\tau=0.478$). The ordering indicates that backbone choice materially impacts evaluation reliability, and the higher $\tau$ for \emph{gemini-2.5-pro} suggests superior preservation of human-preferred rankings. While Pearson values exceed the corresponding Kendall coefficients - as expected given their different sensitivities - the relative gaps are stable across models, reinforcing the robustness of the observed ranking.

\end{document}